\documentclass{article}
\usepackage{spconf,amsmath,amssymb,graphicx}
\usepackage[hidelinks]{hyperref}

\title{POI-Loc: A Fine-Grained POI Localization Benchmark and an
Asymmetric Global-to-Local Matching Method}

\name{Lu Han$^{*}$, Xiting Sun$^{*}$, Hao Wang, Zhiqiang Cao, Ruihuan Du$^{\dagger}$, Ziquan Zeng$^{\dagger}$, Chunlong Lv\thanks{\fontsize{9}{11}\selectfont $^{*}$Equal contribution. $^{\dagger}$Corresponding authors.}}
\address{Amap, Alibaba Group, Beijing, China}

\begin{document}
\ninept
\maketitle
\begin{abstract}
Point-of-interest (POI) localization matches user-provided storefront close-ups to the same shops in wide, geo-tagged vehicle-mounted street views. POIs may change while the surrounding scene stays similar, so scene-level recognition alone cannot establish POI identity. Differences in target scale and capture domains further challenge matching. We introduce POI-Loc, to our knowledge the first benchmark dedicated to this asymmetric, fine-grained POI localization task. Many visual place recognition methods represent each image with a single global vector, which tends to dilute fine-grained features of small storefronts amid background clutter. We propose GLAM (Global-to-Local Asymmetric Matching) to combine global and local evidence. In stage one, a single attention-pooled query probe is matched against compact reference region tokens via learnable soft top-k interaction, with the resulting local similarity fused with global similarity for retrieval. Stage two reuses query region tokens before attention pooling and stored reference tokens for mutual-nearest-neighbor re-ranking. GLAM surpasses both global and two-stage baselines on Recall@1/5/10 and mAP, with about $5\times$ smaller re-ranking features and $280\times$ lower per-pair matching cost than FoL. The benchmark and code will be released at \url{https://github.com/roadhan/glam}.
\end{abstract}

\begin{keywords}
POI localization, visual place recognition,  image retrieval, benchmark, late interaction
\end{keywords}
\section{Introduction}
\label{sec:intro}

A key task in map services is point-of-interest (POI) localization. Given a user-provided close-up of a storefront, signboard, or building entrance, the system retrieves images of the same POI from a large-scale geo-tagged vehicle-mounted street-view database. This supports map construction and updating, POI verification, and location-based services. Moreover, as camera-equipped intelligent vehicles achieve ever-broader road coverage, maintaining POI status directly from vehicle-mounted imagery is drawing increasing attention.

\begin{figure}[t!]
\centering
\includegraphics[width=\columnwidth]{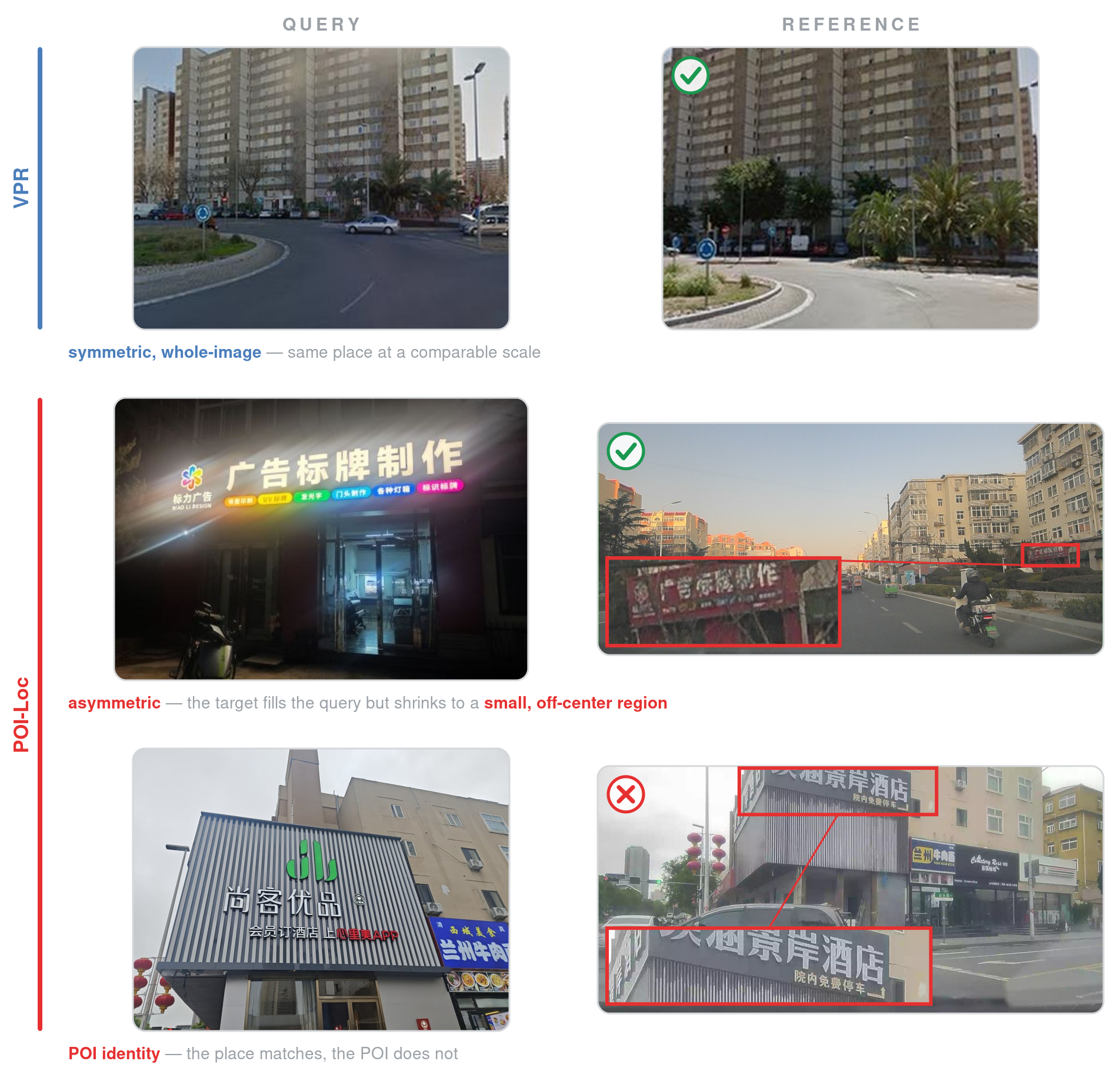}
\caption{place-level VPR versus POI localization in POI-Loc. Marks denote ground-truth
correspondence. Top: a GSV-Cities example, where query and
reference cover the same place at a comparable scale. Middle: in POI-Loc the target
storefront fills the close-up query but shrinks to a small, off-center region of the wide
vehicle-mounted reference (red boxes; magnified insets). Bottom: the reference depicts
the same street corner as the query but a \emph{different tenant}---the place matches,
the POI does not.}
\label{fig:intro}
\end{figure}

Visual place recognition (VPR) retrieves images of the same place~\cite{survey1,survey2}. Benchmarks such as Pittsburgh~\cite{netvlad}, Tokyo~24/7~\cite{tokyo}, SF-XL~\cite{cosplace}, and GSV-Cities~\cite{gsvcities} have advanced scene-level recognition, including cross-domain matching and generalization to unseen places~\cite{nordland,sped,robotcar,msls}. However, a scene may contain multiple POIs, and its shops may open, close, or be replaced while the surrounding streetscape remains similar (Fig.~\ref{fig:intro}, bottom). Recognizing the same scene therefore does not establish the identity of a specific POI. POI-Loc pairs user-contributed close-ups with independently collected vehicle-mounted street views. Its annotations distinguish the queried POI from neighboring shops and previous tenants at the same location.

This POI-level task presents three challenges in POI-Loc (Fig.~\ref{fig:intro}). (i)~Scale asymmetry---a small target in a large scene: the target storefront fills most of the query frame, yet occupies only a small, often off-center region of the reference, competing with neighboring shops, signboards, and background clutter. (ii)~Fine-grained entity matching: visually similar neighboring shops and successive tenants at the same location make scene similarity insufficient to establish POI identity. (iii)~Capture-domain heterogeneity: query and reference are produced by different devices for different purposes, inducing a persistent domain gap. A single global descriptor summarizes the entire reference and can dilute the small target amid surrounding clutter, making fine-grained POI discrimination difficult.

Many VPR methods encode each image as a single global descriptor, which can dilute small storefront regions amid background clutter. NetVLAD~\cite{netvlad} and GeM~\cite{gem} aggregate local features through pooling, while CosPlace~\cite{cosplace} and EigenPlaces~\cite{eigenplaces} use classification-based training to learn retrieval descriptors. MixVPR~\cite{mixvpr} and Conv-AP~\cite{gsvcities} develop alternative feature aggregation designs. More recent methods combine DINOv2~\cite{dinov2} features with aggregators such as SALAD~\cite{salad}, BoQ~\cite{boq}, and ImAge~\cite{image}.

Two-stage methods, including Patch-NetVLAD~\cite{patchnetvlad}, TransVPR~\cite{transvpr}, R2Former~\cite{r2former}, and DELG~\cite{delg}, use local features to refine retrieved candidates. FoL~\cite{fol}, for example, uses mutual-nearest-neighbor matching for re-ranking. These comparisons provide finer matching evidence but add computation and feature storage. For dense pairwise matching, the cost grows with the product of the query and reference token counts. Re-ranking can only refine the selected candidates. GLAM therefore uses local evidence during initial retrieval, before candidate selection.

We propose GLAM (Global-to-Local Asymmetric Matching), which combines a global retrieval anchor with local matching inspired by late interaction~\cite{colbert}. Unlike the symmetric comparison of whole-image vectors in global-descriptor VPR, GLAM performs asymmetric local matching in its first stage: a single attention-pooled query probe searches reference region tokens through soft top-$k$ interaction. A single probe summarizes the target-focused query, while reference tokens preserve the regions to be searched. The local and global scores are fused for retrieval. In the second stage, query region tokens before attention pooling and stored reference tokens are reused for mutual-nearest-neighbor re-ranking. The compact tokens reduce the storage and matching cost of this second stage.

The main contributions are:
\begin{itemize}\setlength{\itemsep}{1pt}
\item We formalize entity-level POI retrieval and introduce POI-Loc, to our knowledge the first benchmark dedicated to this asymmetric, fine-grained task.
\item We propose GLAM, which combines global retrieval with asymmetric local matching and reuses region tokens for lightweight re-ranking.
\item Experiments on POI-Loc show that GLAM outperforms the evaluated baselines on Recall@1/5/10 and mAP, with lower re-ranking feature storage and matching cost than FoL.
\end{itemize}

\section{POI-Loc Benchmark}
\label{sec:benchmark}

\subsection{Data Collection and Curation}
\label{ssec:data}

POI-Loc is built from two independent production streams covering POIs in urban areas of Shenzhen and Qingdao, China. Queries are user-contributed close-ups of storefronts, signboards, or building entrances. They vary in framing and aspect ratio, with the target dominating the image; location metadata may be missing or unreliable. References come from vehicle-mounted cameras that continuously capture wide-field views of urban roads. They cover varied weather and lighting conditions, with motion blur, windshield reflections, occlusions, and differences in viewpoint and image quality.

During dataset construction, we first select POIs and their associated query images from the two cities. The coordinates of each selected POI are then used to select nearby candidate frames from the street-view collection. Trained annotators label the query--candidate pairs according to whether they depict the same POI. References depicting neighboring shops or previous tenants at the same location instead of the queried POI are marked as hard negatives.

\subsection{Overview and Problem Formulation}
\label{ssec:formulation}

POI-Loc contains 1,215 POIs and 11,133 reference images, with a 972-POI training pool and 243 test POIs, disjoint at both the POI and image levels. Under this open-set protocol, all test POIs are unseen during training. The training pool is partitioned into 872 training and 100 validation POIs, with validation used exclusively for checkpoint and hyperparameter selection.

Given a user-uploaded close-up query $q_i$ and a vehicle-mounted reference image $r_j$, we learn a correspondence score $s_{ij}=f_\theta(q_i,r_j)$. The target relation is $y_{ij}=1$ if $r_j$ depicts the queried POI and $0$ otherwise; scene similarity alone does not establish a positive match. The objective is to rank positive references ahead of negatives in the gallery $\mathcal{D}=\bigcup_i\mathcal{G}_i$, where $\mathcal{G}_i$ contains the spatially neighboring frames collected as annotation candidates for POI $i$. The test gallery pools candidates from all 243 test POIs into 2,351 distinct references. We report Recall@1/5/10 and mAP over the 170 queries with annotated positives; the remaining 73 POIs contribute only distractors.

\section{Method: GLAM}
\label{sec:method}

\begin{figure*}[t]
\centering
\includegraphics[width=0.68\textwidth]{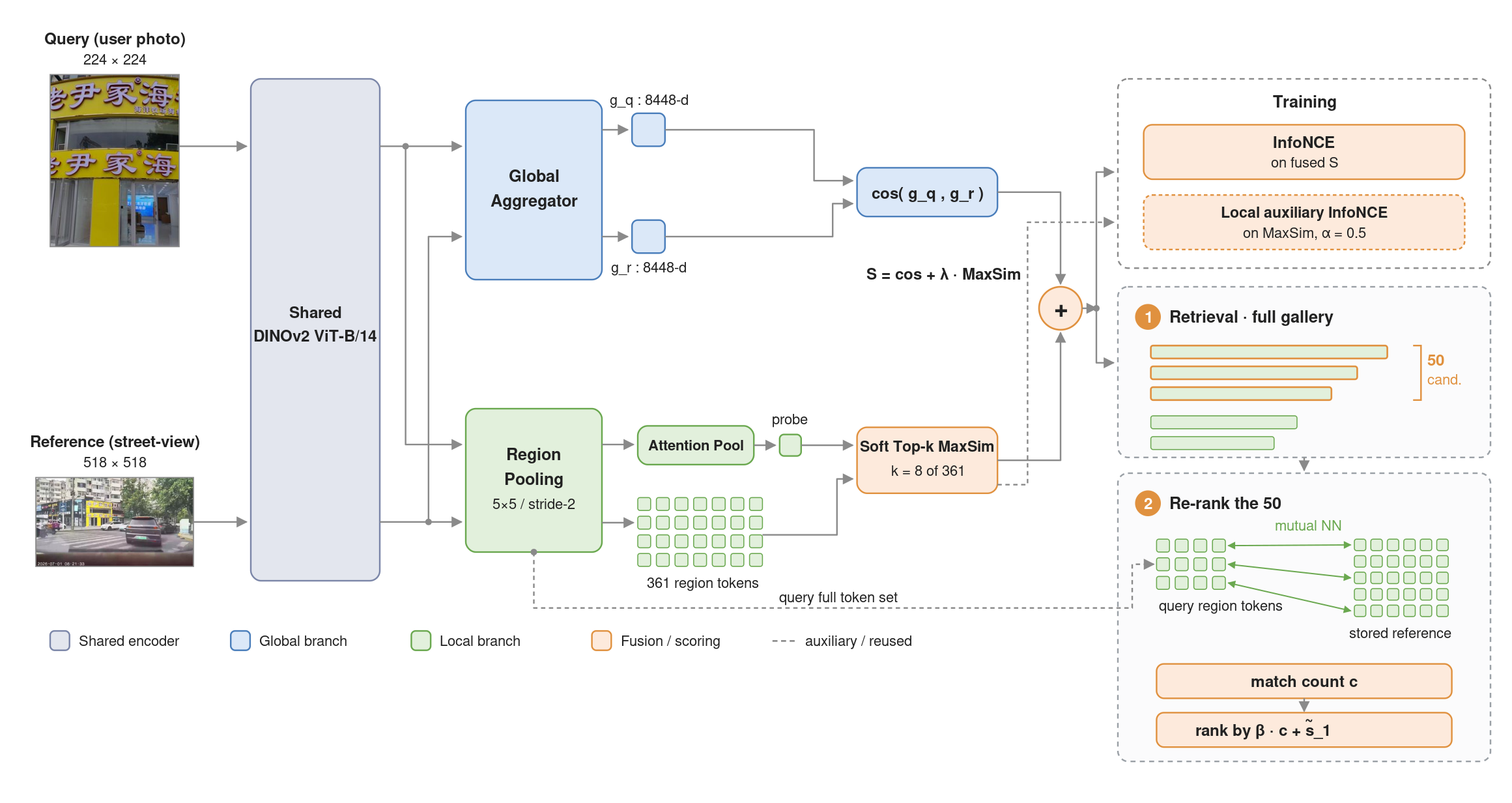}
\caption{Overview of GLAM. A shared DINOv2 encoder feeds a SALAD-based global retrieval anchor~\cite{salad} and an asymmetric local branch. The latter matches a single attention-pooled query probe against reference region tokens via soft top-$k$ MaxSim. Global and local scores are fused through a learnable $\lambda$ for first-stage retrieval, with InfoNCE supervision on both fused and local scores. Candidate re-ranking reuses the query region tokens before attention pooling and the stored reference tokens, combining mutual-nearest-neighbor match counts with the rescaled first-stage score.}
\label{fig:method}
\end{figure*}

GLAM (Fig.~\ref{fig:method}) uses a shared encoder with two branches: a global retrieval anchor and an asymmetric local pathway. The global branch provides image-level similarity, while the local branch matches a single query probe against reference region tokens to capture fine-grained POI evidence. Their scores are fused for first-stage retrieval. The second stage reuses region tokens from both images for lightweight re-ranking.

\subsection{Shared Encoder and Global Anchor}
\label{ssec:global}

Both sides pass through a shared DINOv2 ViT-B/14 backbone~\cite{dinov2} whose last two transformer blocks are fine-tuned, yielding a grid of patch features and a class token. The global branch applies a SALAD-style optimal-transport aggregator with a dustbin~\cite{salad}, producing an 8448-d $L_2$-normalized descriptor per image; global similarity is their cosine. We keep this branch not merely as a strong baseline. Because query and reference traverse the same transformation, the branch inherits the backbone's pretrained similarity structure and retrieves sensibly from the very first epoch, giving the whole model a non-zero starting point; throughout training it acts as a low-variance anchor that regularizes the otherwise unstable local pathway. 

\subsection{Asymmetric Local Pathway}
\label{ssec:local}

The local pathway must answer a question the global branch cannot: \emph{where} in the wide reference is the query's storefront? From the same patch grid we build region descriptors using parameter-free $5{\times}5$ average pooling with stride 2 and padding 2, followed by a learnable $1{\times}1$ projection to $D{=}256$ and $L_2$ normalization. A reference at $518^2$ has a $37{\times}37$ patch grid, yielding $19{\times}19{=}361$ region tokens. The query at $224^2$ passes through the same operations, yielding $8{\times}8{=}64$ region tokens. Attention pooling uses a learned query vector to attend over these tokens, producing a single probe that is then $L_2$-normalized. Two decisions matter here. First, the spatial pooling itself has no learnable parameters: average pooling yields higher re-ranked retrieval scores than learnable region convolutions, including a stride-1 variant that retains $3.8{\times}$ more tokens per reference (Section~\ref{ssec:ablation}). Second, the read-out is intentionally asymmetric---probe for the query, token set for the reference---mirroring the structure of the task: the query does not need to be searched, the reference does.

\subsection{Fused Late-Interaction Scoring}
\label{ssec:fusion}

The local score is a soft top-$k$ MaxSim~\cite{colbert}: with cosine similarities $s_j = \langle \mathbf{p}, \mathbf{r}_j \rangle$ between the probe $\mathbf{p}$ and the reference tokens $\{\mathbf{r}_j\}$, the $k$ ($k{=}8$) largest values are aggregated by a softmax-weighted average, $\mathrm{MaxSim}_k(\mathbf{p}, \{\mathbf{r}_j\}) = \sum_{j \in \mathcal{T}_k} w_j\, s_j$ with $w_j = \exp(\gamma s_j) / \sum_{j' \in \mathcal{T}_k} \exp(\gamma s_{j'})$, where $\mathcal{T}_k$ indexes the top-$k$ similarities and $\gamma$ is a learnable inverse temperature. This soft selection reduces to a hard maximum for $k{=}1$ and is markedly more stable than a hard maximum over hundreds of regions. The final similarity fuses the two branches,
\begin{equation}
\label{eq:fused}
S(q, r) = \langle \mathbf{g}_q, \mathbf{g}_r \rangle + \lambda \cdot \mathrm{MaxSim}_k\big(\mathbf{p}, \{\mathbf{r}_j\}\big),
\end{equation}
with $\lambda = \mathrm{softplus}(\tilde{\lambda})$ learnable. We initialize $\lambda$ small so that the global anchor dominates early training; as the region tokens become discriminative, the model raises $\lambda$ on its own.

\subsection{Training Objective}
\label{ssec:objective}

We train with an in-batch InfoNCE over the fused similarity matrix, treating each query's annotated reference as the positive and masking other same-place pairs:
\begin{equation}
\label{eq:loss}
\mathcal{L} = -\frac{1}{B}\sum_{i=1}^{B}\log\frac{\exp\big(S(q_i, r_i)/\tau_s\big)}{\sum_{j\in\mathcal{A}_i}\exp\big(S(q_i, r_j)/\tau_s\big)} + \alpha\,\mathcal{L}_{\mathrm{loc}},
\end{equation}
where $B$ is the batch size and $\tau_s$ a learnable temperature. The valid index set $\mathcal{A}_i$ retains the paired positive $i$ and excludes other batch entries with the same POI identity. The auxiliary loss $\mathcal{L}_{\mathrm{loc}}$ applies the same InfoNCE and mask to the local-only score $\mathrm{MaxSim}_k(\mathbf{p}, \{\mathbf{r}_j\})$, with weight $\alpha{=}0.5$. It directly supervises local matching, encouraging region tokens to remain useful independently of the global score.

\subsection{Lightweight Region-Token Re-ranking}
\label{ssec:rerank}

GLAM reuses existing region descriptors without training a separate re-ranking model. After stage-1 fused retrieval, the top-$N$ candidates ($N{=}50$) undergo mutual-nearest-neighbor matching with cosine threshold $\tau{=}0.7$. Matching uses the query region tokens before attention pooling and each candidate's stored reference tokens, rather than the single stage-1 probe. Integer match counts can yield ties, so we combine them with first-stage retrieval evidence. The selected candidates remain above the rest of the gallery and are ranked by $\beta\,c + \tilde{s}_1$, where $c$ is the match count, $\tilde{s}_1 \in [0,1)$ is the stage-1 score rescaled within the candidate set, and $\beta$ controls the contribution of local matches. This joint score balances both sources of evidence and uses the stage-1 score to distinguish candidates with equal counts. Re-ranking adds local matching computation but requires no separate feature extraction (Section~\ref{ssec:eff-exp}).
\section{Experiments}
\label{sec:exp}

\subsection{Setup}
\label{ssec:setup}

All methods share the same backbone (DINOv2 ViT-B/14, last two transformer blocks fine-tuned) and are trained and evaluated under the protocol of Section~\ref{ssec:formulation}. We use batch size 64 and AdamW with learning rate $10^{-4}$ and weight decay $10^{-4}$. The learning rate follows a 10-epoch linear warmup, with decay scheduled at epoch 40. Training runs for at most 60 epochs, with early stopping based on validation R@1 and a patience of 10 epochs. For each run, we select one checkpoint by validation R@1 and report all four metrics from it on the test split. The fusion weight $\beta$ (Section~\ref{ssec:rerank}) is fixed the same way, by lexicographic order on validation (R@1, R@5, R@10), giving $\beta{=}0.5$; the sweep covers $\beta{>}0$, as $\beta{=}0$ recovers stage-1 exactly. All evaluations are conducted on POI-Loc to assess POI-level correspondence from close-up queries to wide street views, rather than scene-level place recognition.

\subsection{Main Results}
\label{ssec:main}

\begin{table}[t]
\caption{Main results on POI-Loc (open-set test set: 170 queries against 2,351 references); all methods follow the protocol of Section~\ref{ssec:setup}.}
\label{tab:main}
\centering
\fontsize{9}{11}\selectfont
\setlength{\tabcolsep}{3pt}
\begin{tabular}{lcccc}
\hline
Method & R@1 & R@5 & R@10 & mAP \\
\hline
\multicolumn{5}{l}{\emph{Global descriptor (one vector per image)}}\\
CosPlace~\cite{cosplace} & 1.2 & 4.1 & 7.6 & 2.4 \\
BoQ~\cite{boq} & 8.8 & 17.6 & 22.9 & 10.7 \\
EDTformer~\cite{edtformer} & 8.8 & 22.4 & 28.2 & 11.9 \\
ImAge~\cite{image} & 10.6 & 21.2 & 28.2 & 12.3 \\
SALAD~\cite{salad} & 11.8 & 25.3 & 31.2 & 14.2 \\
\hline
\multicolumn{5}{l}{\emph{Two-stage with local re-ranking}}\\
FoL (stage-1)~\cite{fol} & 8.2 & 19.4 & 22.9 & 11.7 \\
FoL (+re-ranking)~\cite{fol} & 12.9 & 21.8 & 26.5 & 13.7 \\
SelaVPR (stage-1)~\cite{selavpr} & 11.2 & 20.6 & 29.4 & 12.8 \\
SelaVPR (+re-ranking)~\cite{selavpr} & 12.9 & 22.9 & 30.0 & 14.2 \\
\hline
\multicolumn{5}{l}{\emph{Ours}}\\
\textbf{GLAM (fused, stage-1)} & 13.3 & 30.0 & 36.5 & 16.6 \\
\textbf{GLAM (+re-ranking, $\beta{=}0.5$)} & \textbf{18.2} & \textbf{33.1} & \textbf{40.0} & \textbf{20.5} \\
\hline
\end{tabular}
\end{table}

Table~\ref{tab:main} summarizes the comparison. GLAM results in this table are averaged over trials with different random seeds. Three observations stand out. First, GLAM's fused stage-1 outperforms all global-descriptor baselines and both FoL and SelaVPR two-stage pipelines on all four metrics, supporting the benefit of local evidence before candidate selection. Second, local re-ranking improves SelaVPR from $11.2$ to $12.9$ R@1 and from $12.8$ to $14.2$ mAP; GLAM reaches $18.2$ R@1 and $20.5$ mAP, outperforming both two-stage baselines. Third, no method exceeds $20$ R@1: this setting is far from solved.

\subsection{Ablations}
\label{ssec:ablation}

Results in (a) and (b) are averaged over three random seeds after re-ranking; (c) compares branch scores before re-ranking using a single GLAM checkpoint.

\begin{table}[t]
\caption{Ablation on the region operator of the local branch (stage-2). ``Tok.'' is the number of region tokens stored per reference.}
\label{tab:ablation}
\centering
\fontsize{9}{11}\selectfont
\setlength{\tabcolsep}{3pt}
\begin{tabular}{lccccc}
\hline
Region operator & Tok. & R@1 & R@5 & R@10 & mAP \\
\hline
Learn.\ conv., $3{\times}3$, stride 1 & 1369 & 12.4 & 28.6 & 37.5 & 16.2 \\
Learn.\ conv., $5{\times}5$, stride 2 & 361 & 11.8 & 25.7 & 34.9 & 14.8 \\
\textbf{Avg.\ pooling, $5{\times}5$, stride 2} & \textbf{361} & \textbf{18.2} & \textbf{33.1} & \textbf{40.0} & \textbf{20.5} \\
\hline
\end{tabular}
\end{table}

\textbf{(a) Region operator.}
Table~\ref{tab:ablation} refines the design choice of Section~\ref{ssec:local}. At an equal token budget (361), parameter-free averaging outperforms the learnable convolution on all four re-ranked metrics. The stride-1 convolution also remains below averaging on every metric despite storing $3.8\times$ more tokens per reference and incurring approximately $15\times$ the pairwise re-ranking matching cost. We therefore adopt average pooling.

\begin{table}[t]
\caption{Ablation on the number of regions $k$ in the soft top-$k$ MaxSim (stage-2).}
\label{tab:topk}
\centering
\fontsize{9}{11}\selectfont
\setlength{\tabcolsep}{4pt}
\begin{tabular}{lcccc}
\hline
$k$ & R@1 & R@5 & R@10 & mAP \\
\hline
$k=1$ (hard maximum) & 16.7 & 28.0 & 37.6 & 17.9 \\
$\mathbf{k=8}$ \textbf{(ours)} & \textbf{18.2} & \textbf{33.1} & 40.0 & \textbf{20.5} \\
$k=16$ & 16.9 & 32.0 & \textbf{41.0} & 19.5 \\
\hline
\end{tabular}
\end{table}

\noindent\textbf{(b) Soft top-$k$ MaxSim.}
Both soft top-$k$ settings improve all four re-ranked metrics over the hard maximum ($k{=}1$) in Table~\ref{tab:topk}. The default $k{=}8$ gives the highest mean R@1, R@5, and mAP, while $k{=}16$ achieves a higher R@10. Thus, increasing $k$ beyond 8 does not consistently improve final retrieval performance.

\noindent\textbf{(c) The local branch is discriminative on its own.}
The auxiliary term of Eq.~(\ref{eq:loss}) turns the region tokens into stand-alone features: the local score alone retrieves at 9.4 R@1 / 28.8 R@10 against 11.2 / 34.7 for the global branch, and fusing them ($\lambda{=}0.32$) yields 12.9 / 36.5. This is what makes the re-ranking stage of Section~\ref{ssec:rerank} free.

\subsection{Efficiency}
\label{ssec:eff-exp}

The asymmetric re-ranking design of Section~\ref{ssec:rerank} is also markedly cheaper than dense local re-ranking. GLAM keeps $361{\times}256$ region tokens per reference (0.18 MB in fp16) against FoL's $3600{\times}128$ (0.92 MB), a $5\times$ reduction in stored local data, and because only $64$ query region tokens are matched against them, one query--reference comparison costs ${\approx}5.9$ M-MACs versus ${\approx}1.66$ G-MACs for FoL---about two orders of magnitude less---while delivering higher accuracy (Table~\ref{tab:main}).

\subsection{Qualitative Analysis}
\label{ssec:qualitative}

\begin{figure}[t]
\centering
\includegraphics[width=\columnwidth]{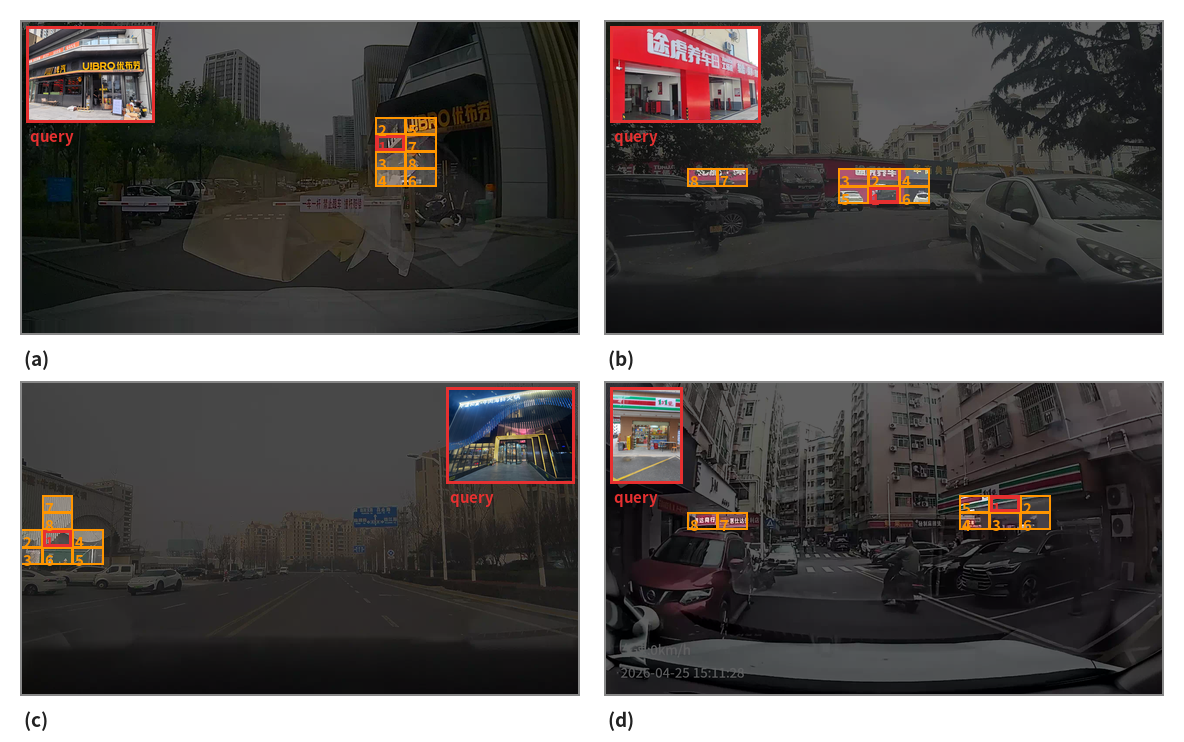}
\caption{Top-8 region tokens (numbered boxes; 1 = largest weight) selected by the soft top-$k$ MaxSim when matching each close-up query (red inset) against the dimmed reference, under (a)~strong windshield reflections, (b)~occlusion by parked vehicles, (c)~a day--night gap, and (d)~dense street clutter.}
\label{fig:heatmap}
\end{figure}

Although GLAM is trained with retrieval supervision only, the tokens selected by the soft top-$k$ MaxSim concentrate on the queried storefront and form spatially contiguous clusters (Fig.~\ref{fig:heatmap}), providing direct evidence that the asymmetric probe-to-token interaction performs implicit localization inside the wide reference without any spatial supervision.

\section{Conclusion}
\label{sec:conclusion}

We studied point-of-interest localization as an entity-level retrieval problem in which a close-up storefront query must be matched to the same POI embedded in wide vehicle-mounted street views, and introduced POI-Loc, to our knowledge the first benchmark dedicated to this asymmetric, fine-grained POI localization task under an open-set protocol. On POI-Loc, global-descriptor VPR methods are systematically limited, whereas the proposed GLAM---a global anchor coupled with an asymmetric local pathway whose region tokens double as a lightweight mutual-nearest-neighbor re-ranker---achieves the best accuracy with lower re-ranking storage and matching cost than FoL. Future work includes scaling POI-Loc and exploring richer, yet still asymmetric and index-friendly, local interactions.

\vfill\pagebreak

\bibliographystyle{IEEEbib}
\bibliography{refs}

\end{document}